\documentclass[conference]{IEEEtran}

\usepackage{amsmath,amssymb,amsfonts}
\usepackage{algorithmic}
\usepackage{graphicx}
\usepackage{textcomp}
\usepackage{xcolor}
\usepackage{natbib}
\setcitestyle{numbers,square}
\usepackage{pifont}

\usepackage{comment}
\usepackage{wrapfig}
\usepackage{colortbl}
\usepackage{listings}
\usepackage{hyperref}
\hypersetup{colorlinks,allcolors=blue}
\usepackage{amssymb}
\usepackage{subcaption}
\usepackage{pifont}
\usepackage{tikz}
\usetikzlibrary{math}
\usepackage{threeparttable}
\usepackage{booktabs}
\usepackage{multirow}
\usepackage{graphicx} 
\usepackage{float} 
\usepackage[utf8]{inputenc} 
\usepackage[T1]{fontenc}    
\usepackage{hyperref}       
\usepackage{url}            
\usepackage{booktabs}       
\usepackage{amsfonts}       
\usepackage{nicefrac}       
\usepackage{microtype}      
\usepackage{xcolor}         
\usepackage{rotating}

\definecolor{dkgreen}{rgb}{0,0.6,0}
\definecolor{gray}{rgb}{0.5,0.5,0.5}
\definecolor{mauve}{rgb}{0.58,0,0.82}
\definecolor{almond}{rgb}{0.94, 0.87, 0.8}
\definecolor{babyblueeyes}{rgb}{0.63, 0.79, 0.95}
\definecolor{beige}{rgb}{0.96, 0.96, 0.86}
\definecolor{anti-flashwhite}{rgb}{0.95, 0.95, 0.96}
\definecolor{darkblue}{rgb}{0.03, 0.27, 0.49}
\definecolor{darkgreen}{rgb}{0.01, 0.75, 0.24}
\definecolor{darkred}{rgb}{0.76, 0.23, 0.13}
\definecolor{light-gray}{gray}{0.92}

\def\BibTeX{{\rm B\kern-.05em{\sc i\kern-.025em b}\kern-.08em
    T\kern-.1667em\lower.7ex\hbox{E}\kern-.125emX}}
    
\begin{document}

\title{Camera-Based HRV Prediction for Remote Learning Environments
}

\author{
    \IEEEauthorblockN{Kegang Wang$^{a,c}$, Yantao Wei$^{a*}$, Jiankai Tang$^{b,c}$, Yuntao Wang$^{b,c*}$
    \IEEEauthorblockN{Mingwen Tong$^{a}$, Jie Gao$^{a}$, Yujian Ma$^{d}$, Zhongjin Zhao$^{a}$}}
    \IEEEauthorblockA{\small $^*$ Co-Corresponding author}
    \IEEEauthorblockA{\small $^a$ Central China Normal University, Wuhan, China, 430079}
    \IEEEauthorblockA{\small $^b$ National Key Laboratory of Human Factors Engineering, Beijing, 100094. }
    \IEEEauthorblockA{\small $^c$ Department of Computer Science and Technology, Tsinghua University, Beijing, 100084.}
    \IEEEauthorblockA{\small $^d$ East China Normal University, Shanghai, China, 200241}
}
\maketitle

\begin{abstract}

In recent years, due to the widespread use of internet videos, remote photoplethysmography (rPPG) has gained more and more attention in the fields of affective computing. Restoring blood volume pulse (BVP) signals from facial videos is a challenging task that involves a series of preprocessing, image algorithms, and postprocessing to restore waveforms. 
Not only is the heart rate metric utilized for affective computing, but the heart rate variability (HRV) metric is even more significant. 
The challenge in obtaining HRV indices through rPPG lies in the necessity for algorithms to precisely predict the BVP peak positions. 
In this paper, we collected the Remote Learning Affect and Physiology (RLAP) dataset, which includes over 32 hours of highly synchronized video and labels from 58 subjects. This is a public dataset whose BVP labels have been meticulously designed to better suit the training of HRV models. Using the RLAP dataset, we trained a new model called Seq-rPPG, it is a model based on one-dimensional convolution, and experimental results reveal that this structure is more suitable for handling HRV tasks, which outperformed all other baselines in HRV performance and also demonstrated significant advantages in computational efficiency.

\end{abstract}

\begin{IEEEkeywords}
remote photoplethysmography, dataset, affective computing, remote learning.
\end{IEEEkeywords}

\section{Introduction}

In recent years, many publicly available datasets have emerged in the field of remote physiological sensing\citep{afrl}, \citep{mmsehr}, \citep{viplhr}, \citep{pure}, \citep{ubfcphys}, \citep{ubfcrppg}, \citep{obf}, \citep{mahnob}, \citep{mcduff2022scamps}, \citep{tang2023mmpd}, \citep{mrnird}, \citep{mrnird1}, \citep{vicarppg}, with numerous datasets focusing on providing benchmark tests for scenarios with greater diversity in motion, age, ethnicity, gender, and lighting conditions. 



However, most datasets employ two separate devices to collect video and physiological signal labels, lacking mechanisms to ensure strict synchronization between these two signals. Upon careful examination, it was found that some data were not precisely synchronized; others exhibited frame rate fluctuations leading to delays starting from a certain point in time; and some datasets utilized manual coarse synchronization, still displaying errors ranging from 100 to 200 milliseconds. Typically, the time scale of HRV analysis is several tens of milliseconds, and the presence of these errors makes the datasets less suitable for HRV training. Moreover, HRV requires purer BVP signals; therefore, it is equally crucial that the datasets possess high video quality, McDuff et al.(2017)\citep{compression} and Yu et al.(2019)\citep{steven} have indicated that compressed video formats can degrade BVP signals, hence the importance of collecting lossless format data.

Prior research\citep{physnet}, \citep{talos}, \citep{sun2022byhe} has indicated that problems related to synchronization make models more challenging to train, particularly evident when Mean Squared Error (MSE) is employed as the loss function. Since MSE is overly sensitive to delays, a range of loss functions either insensitive to delays or invariant to them have been proposed\citep{physnet}, \citep{talos}, \citep{sun2022byhe}, \citep{rtrppg}, \citep{physformer}. Nevertheless, there remains a scarcity of datasets designed to address this issue, specifically datasets with high degrees of synchronization, which continue to be challenging to compile.


Building upon previous work, we have collected the Remote Learning Affect and Physiology (RLAP) dataset for use in remote or online learning contexts. This dataset includes high-quality video and highly synchronized BVP labels. Our goal is to enhance HRV accuracy through high-quality data, and to expand the application of rPPG in affect computing\citep{greene2016survey}, particularly in the emotional analysis of students, and can achieve higher accuracy Interbeat Interval (IBI), extend to LLM-based health models\citep{tang2023alpha}. Basic information about RLAP, and comparisons with other datasets, can be found in Table \ref{dataset}.

\begin{table*}[!htp]
\centering
 \caption{Basic information of major datasets and our new dataset RLAP}
 \label{dataset}
  \scalebox{1.3}{
 \begin{threeparttable}
\begin{tabular}{ccccccc}
   \toprule
   Dataset &Participants &Frames& Hours & PPG & Signal offset & Lossless format \\
   \midrule
   AFRL\citep{afrl}\textcolor{red}{$^{1}$} & 25 & 97.2M\textcolor{red}{$^{2}$} & 25 & \checkmark & 0 & \checkmark \\
   PURE\citep{pure} & 10 & 106K&1 & \checkmark & 0 & \checkmark \\
   UBFC\citep{ubfcrppg} & 42 & 75K&0.7 & \checkmark & >0.5s\textcolor{red}{$^{3}$} & \checkmark \\
   MMSE-HR\citep{mmsehr} & 58 & 435K&4.8 &  & 0 &  \\
   MAHNOB-HCI\citep{mahnob} & 30 & 25.2M\textcolor{red}{$^{4}$}&19.4 &  & 0 &  \\
   VIPL-HR\citep{viplhr} & 107 & 2.14M\textcolor{red}{$^{5}$}&19.8 & \checkmark & >0.5s & \\
   MMPD\citep{tang2023mmpd} & 33 & 1.15M&10.6 & \checkmark & <0.2s & \\
   \rowcolor{light-gray} RLAP &58& 3.53M &32.7 & \checkmark & 0 & \checkmark \\
   \bottomrule
\end{tabular}

 \begin{tablenotes}[flushleft]
 \item[1] The authors did not specify that this is a public dataset.
 \item[2] Uses nine RGB cameras to synchronously record at 120 fps.
 \item[3] Only a portion of the videos have significant offsets.
 \item[4] Uses one RGB camera and five BW cameras to synchronously record at 60 fps.
  \item[5] Uses an estimated 30 fps even though the actual fps fluctuates between 15 and 30.
 \end{tablenotes}
 \end{threeparttable}
 }
 \end{table*}



Mobile-device-based local rPPG algorithms are particularly promising since they imply reduced computational costs and thorough protection of privacy, prompting numerous works\citep{physnet}, \citep{deepphys}, \citep{mttscan}, \citep{rtrppg}, \citep{talos} to concentrate on lightweight end-to-end networks. In this paper, we design an algorithm based on a one-dimensional convolutional neural network (1D CNN) that encodes video frames into one-dimensional features via straightforward linear mapping, and extracts BVP signals. Our experiments show that this method not only offers superior HRV accuracy but also reduces computational overhead significantly.

This paper makes the following contributions:
\begin{itemize}



\item A high-quality, highly synchronized public dataset, RLAP, has been constructed, designed for emotional scenes in remote learning contexts. As an additional contribution, PhysRecorder, the tool we developed for collecting this dataset, has also been open-sourced. 

\item A lightweight algorithm based on 1D CNN, Seq-rPPG, has been proposed, demonstrating significant advantages in heart rate variability tasks with lower computational overhead.

\end{itemize}

\section{Dataset}

The collection of the RLAP dataset involves two steps: 1) Examination of the camera's supported codecs and transmission standards to ensure the most lossless storage format possible; examination of the pulse oximeter's API to ensure real-time acquisition of the BVP signal; and development of software that simultaneously collects these two types of signals. 2) Following the pre-defined data collection procedures, set up the collection environment, communicate with each participant, and collect data after obtaining signed informed consent.

\begin{figure*}[!htb] 
\centering 
\includegraphics[width=1\textwidth]{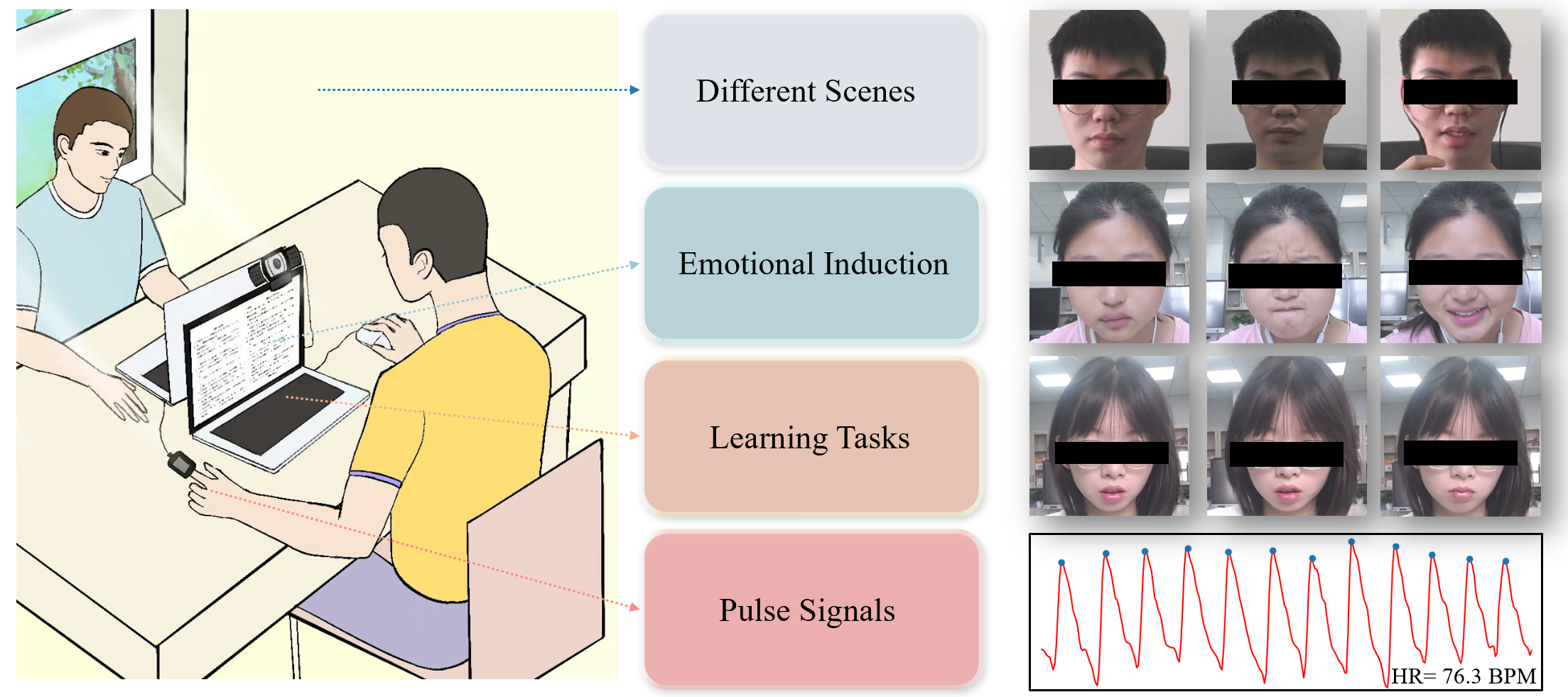} 
\caption{Overview of the RLAP dataset. The RLAP dataset comprises 58 student samples, encompassing various scenarios, emotions, and levels of learning engagement. While completing these tasks, participants' pulse signals were synchronously recorded with a pulse oximeter.} 
\label{rlap} 
\end{figure*}

\begin{table*}
\centering
\caption{Data collection workflows}
\label{table2}
\scalebox{1.1}{
\begin{threeparttable}
    \begin{tabular}[t]{clcccccc}
    \toprule
    \bf Sub-dataset & \bf Task or scenario & \bf Target & \bf Duration(S) & \bf Camera codec & \bf Video codec & \bf Resolution \\
    \midrule
    \multirow{4}*{Scene tasks} & Relaxed & - & 120& \multirow{4}*{YUY2\textcolor{red}{$^1$}} & \multirow{4}*{RGB,MJPG,H264} & \multirow{4}*{640×480} \\
    ~ & Relaxed (dark) & - & 120 & ~ & ~ & ~  \\
    ~ & Play a game & Move hand & 120 & ~ & ~ & ~  \\
    ~ & Read an article & Facial activities& 120  & ~  & ~ & ~ \\
    \hline
    \multirow{6}*{Emotional tasks} & Natural scenery& Tranquility & 120& \multirow{6}*{MJPG} & \multirow{6}*{MJPG,H264} & \multirow{6}*{1920×1080} \\
    ~ & Puzzle game& Concentration & 180 & ~ & ~ & ~  \\
    ~ & Comedy& Happiness & 120 & ~ & ~ & ~  \\
    ~ & Illusion picture & Confusion & 20  & ~  & ~ & ~ \\
    ~ & Academic paper& Drowsiness & 60  & ~  & ~ & ~ \\
    ~ & Yawning video& Drowsiness & 60  & ~  & ~ & ~ \\
    \hline
    \multirow{3}*{Learning tasks} & Video-based learning & Video engagement & 240 & \multirow{3}*{MJPG} & \multirow{3}*{MJPG,H264} & \multirow{3}*{1920×1080} \\
    ~ & Textbook-based learning & Text engagement & 480 & ~ & ~ & ~  \\
    ~ & Watch a public class & Low engagement & 420 & ~ & ~ & ~  \\
    \bottomrule
    \end{tabular}
    
     \begin{tablenotes}[flushleft]
     \item[1] YUY2 is a RAW transmission format for webcams, it is limited by bandwidth and can only operate at 480p@30fps.
     \end{tablenotes}
     \end{threeparttable}
}
\end{table*}

\subsection{Program Coding}
We used a Logitech C930c webcam to capture videos that support MJPG and YUY2 (YUV422) formats. By default, it used the MJPG format to achieve 1920x1080@30fps (general scenarios) video transmission and specified YUY2 format through API to permit the transmission of raw images at 640x480@30fps (rPPG specialized scenarios).

We employed the HID driver to read raw signals from a pulse oximeter via the USB interface, capturing specifically the BVP segment. Our programmed application concurrently captured signals from the camera and the pulse oximeter, assigning UNIX timestamps to each value or frame to ensure rigorous alignment of data to prevent errors in the positioning of BVP peaks.

\subsection{Dataset Collection}  

During data recording, subjects completed a series of tasks or watch videos. After completing the specified task, the subject rested for a while, and then the experimenter assigned them the next task. All 58 subjects (16 males and 42 females) were Chinese students, mainly master's degree students. 
The tasks assigned to each subject were divided into three parts. The first part was the rPPG task, which included four scenarios: a general relaxation scenario, a dark relaxation scenario with the curtains drawn 
a tense scenario involving playing a time-related game, and a speaking scenario while reading an article by Lu Xun. 
The second part involved emotional induction tasks, requiring the subjects to watch six videos interspersed with brief rest periods: viewing natural landscapes, solving a puzzle game, watching a comedy, viewing hallucinatory images, reading an academic paper, and watching a yawning video. The third part was the learning engagement task, which involved completing three learning activities, each followed by a short rest: learning and answering questions based on a video (simple), learning and answering questions from a text (difficult), and watching an open course lecture (without any exercises).
RLAP provided more than 32 hours (3.53 million frames) of video. More details about RLAP can be found in Table \ref{table2}. The schematic diagram of the data collection workflows and some samples can be found in Fig. \ref{rlap}

The data collection environment faces a window and has indoor artificial light sources. The subject sits in front of a computer, about one meter away from the camera. 
During the video collection process, the subject holds a mouse or pen with their right hand to complete tasks and wears a CMS50E pulse oximeter on their left hand. They are instructed to minimize left-hand movement to ensure stable signal acquisition. The subjects' heads are not fixed and can move naturally. Meanwhile, the examiner used another computer nearby to connect to the participant's webcam and pulse oximeter, overseeing data collection.

\section{Algorithm}

\begin{figure*}[!htb]
\centering
\scalebox{1.2}{
    \begin{tikzpicture}[rotate=0, transform shape]
    \tikzmath{\offset=0; };
    \node[rotate=90] at (-1.5+\offset, 0){RGB sequence};
    \draw[->](-1.2+\offset, 0) -- (-0.7+\offset, 0);
    \node[rotate=90] (box1) at (-0.2+\offset, 0)[draw,minimum width=3cm,minimum height=0.5cm, fill=almond, align=center]{Conv1D\\C=64, K=3, S=3};
    \node[rotate=90] (box2) at (1+\offset, 0)[draw,minimum width=3cm,minimum height=0.5cm, fill=babyblueeyes]{RealFFT1D};
    \node[rotate=90] (box3) at (2+\offset, 0)[draw,minimum width=3cm,minimum height=0.5cm, align=center, fill=babyblueeyes]{Conv1D-BN-ReLU\\C=128, K=5, S=1};
    \node[rotate=90] (box4) at (3+\offset, 0)[draw,minimum width=3cm,minimum height=0.5cm, fill=babyblueeyes]{InvRealFFT1D};
    \node at (3.6+\offset, 0) {$\bigoplus$}; 
    \node[rotate=90] (box5) at (4.4+\offset, 0)[draw,minimum width=3cm,minimum height=0.5cm, align=center, fill=beige]{Conv1D-BN-ReLU\\C=64, K=10, S=1};
    \node[rotate=90] (box6) at (5.6+\offset, 0)[draw,minimum width=3cm,minimum height=0.5cm, fill=babyblueeyes]{RealFFT1D};
    \node[rotate=90] (box7) at (6.6+\offset, 0)[draw,minimum width=3cm,minimum height=0.5cm, align=center, fill=babyblueeyes]{Conv1D-BN-ReLU\\C=128, K=3, S=1};
    \node[rotate=90] (box8) at (7.6+\offset, 0)[draw,minimum width=3cm,minimum height=0.5cm, fill=babyblueeyes]{InvRealFFT1D};
    \node at (8.2+\offset, 0) {$\bigoplus$}; 
    \node[rotate=90] (box9) at (9+\offset, 0)[draw,minimum width=3cm,minimum height=0.5cm, align=center, fill=beige]{Conv1D-BN-ReLU\\C=32, K=5, S=1};
    \node[rotate=90] (box10) at (10.2+\offset, 0)[draw,minimum width=3cm,minimum height=0.5cm, align=center, fill=anti-flashwhite]{Conv1D\\C=1, K=1, S=1};
    \draw[->](box1) -- (box2);
    \draw[->](box2) -- (box3);
    \draw[->](box3) -- (box4);
    \draw[->](box4) -- (box5);
    \draw[->](box5) -- (box6);
    \draw[->](box6) -- (box7);
    \draw[->](box7) -- (box8);
    \draw[->](box8) -- (box9);
    \draw[->](box9) -- (box10);
    \draw[-](0.5+\offset, 0) -- (0.5+\offset, 2);
    \draw[-](0.5+\offset, 2) -- (3.6+\offset, 2);
    \draw[->](3.6+\offset, 2) -- (3.6+\offset, 0.2);
    \draw[-](5.1+\offset, 0) -- (5.1+\offset, 2);
    \draw[-](5.1+\offset, 2) -- (8.2+\offset, 2);
    \draw[->](8.2+\offset, 2) -- (8.2+\offset, 0.2);
    \draw[->](10.65+\offset, 0) -- (11.2+\offset, 0);
    \node[rotate=90] at (10.9+\offset, 0.6){Flatten};
    \foreach \x in {0.1, 0.7, 1.3, 1.9}{
        \draw[color=red, in=180, out=180] (12.5, -1.5+\x) to (11.5, -1.2+\x);
        \draw[color=red, in=270, out=0] (11.5, -1.2+\x) to (11.8, -1.1+\x);
        \draw[color=red, in=180, out=20] (11.8, -1.1+\x) to (12.5, -0.9+\x);
    }
    \draw[color=red, in=180, out=180] (12.5, -0.9+1.9) to (11.5, -0.6+1.9);

    \node[rotate = 0] at (-0.2+\offset, -1.8) {$\underbrace{\hspace{1cm}}$};
    \node[rotate=0] at (-0.2+\offset, -2.1){Encoder};

    \node[rotate = 0] at (5.1+\offset, -1.8) {$\underbrace{\hspace{8.8cm}}$};
    \node[rotate=0] at (5.1+\offset, -2.1){Decoder};
    
    \end{tikzpicture}
    
}
    
\caption{Seq-rPPG network, it consists of an encoder and a decoder. The encoder is a single-layer 1x1 convolution, while the decoder comprises four layers of alternating time-domain and frequency-domain convolutions.}
\label{Fig.model}
\end{figure*}


HRV tasks require accurate peak estimation, a time-sensitive endeavor. Considering the need to capture as many frequency characteristics as possible, the model architecture is designed to be specialized for time-series tasks and should incorporate a large time window. In contrast to most general spatio-temporal modeling approaches such as Differential 2D Convolution (DIFF 2D CNN)\citep{deepphys}, \citep{efficientphys}, \citep{mttscan}, 3D Convolution (3D CNN)\citep{rtrppg}, \citep{physnet}, \citep{compression}, and Spatio-temporal Maps (STMap \& MSTMap)\citep{synrhythm}, \citep{rhythmnet}, \citep{cvd}, our method focuses predominantly on temporal features. This includes utilizing one-dimensional convolution (1D CNN) along the time dimension and ensuring strict alignment between BVP labels and video during the training phase.

Handcrafted algorithms \citep{ICA}, \citep{ICA1}, \citep{chrom}, \citep{pos}, \citep{samc} are based on separate reflection components. Shafer \citep{shafer} assumes that the BVP signal in the RGB image comes through a linear combination of different frequency rays, while skin-reflected light contains a specular reflection component and a diffuse reflection component. In postprocessing, stationary signals and noise are filtered, while periodic signals generated by fluctuations in hemoglobin concentration are passed. Therefore, our algorithm is divided into two parts. The first part combines the RGB channels and frame numbers of the original video ($450 \times 8 \times8 \times3$) and merges the width and height to obtain a $1350 \times 64$ RGB sequence. A convolution kernel with a size of 3 and a stride of 3 is used to mix the RGB sequence, resulting in a coarse signal of $450\times 64$. The second part involves using multiple convolution filters with activation functions to perform nonlinear filtering on the coarse signal, thereby obtaining the BVP signals.
These two components can be viewed as an encoder-decoder, a structure previously utilized in machine translation tasks for the Seq2Seq network\citep{seq2seq}. From our perspective, the rPPG task shares similarities with machine translation, where the rPPG algorithm "translates" videos into BVP signals, hence it is termed Seq-rPPG.

We drew inspiration from the Fast Fourier Convolution (FFC)\citep{FFC}, \citep{ffcse}, which is very effective in processing periodic signals (e.g., audio information). We designed a spectral transformation module and added it to the 1D CNN to enhance its performance. The final model alternates between four temporal domain CNN layers and spectral transformations (see Fig.\ref{Fig.model}).

We implemented the spectral transformation module using a fast Fourier transform (FFT) and an 1D CNN. For a signal $\textbf{Y}\in\mathbb{R}^{N\times C}$, the spectral filtering layer first performs a real fast Fourier transform (RFFT) on each channel, obtains frequency domain signal $\textbf{Y}_{Freq}\in\mathbb{Z}^{\frac{\lfloor N+1\rfloor}{2}\times C}$, decomposes it into a real part $\textbf{Y}_{Real}$ and an imaginary part $\textbf{Y}_{Img}$, and then combines them on the channel as $\textbf{Y}_{Comb}\in\mathbb{R}^{\frac{\lfloor N+1\rfloor}{2}\times 2C}$. Next, a convolution layer re-decomposes the output into real and imaginary parts. The new output is converted to complex numbers and recovered to the time domain signal by inverse real fast Fourier transform (IRFFT). The output signal is mixed with the original signal through a residual connection. Note that the number of channels remains constant throughout the process. 

\section{Experiment and Results}
The experimental platform used was an AMD Ryzen 9 5950X CPU with an Nvidia RTX 3090 GPU and the Windows 11 OS. We selected five baseline models among which the proposed model, PhysNet\citep{physnet}, DeepPhys\citep{deepphys}, EfficientPhys\citep{efficientphys}, and TS-CAN\citep{mttscan} were trained on Tensorflow 2.6. PhysFormer\citep{physformer} was trained on Pytorch 2.0. Though multiple configurations are detailed in the code, TS-CAN and DeepPhys adopt a resolution of 36x36, in accordance with the original text. For pretrained models, we tested their mobile CPU inference performance on a Raspberry Pi 4B (CPU: Cortex-A72 4 cores; OS: Debian 11). 

The Seq-rPPG uses the following training parameters: Adam optimizer, batch size of 32, and Mean Absolute Error (MAE) loss. The parameters for other algorithms are provided as per the original literature.

\subsection{Datasets and Metrics}
In addition to the RLAP dataset, there are two other datasets used for testing: UBFC\citep{ubfcrppg}, which includes 42 video clips from 42 subjects, with each clip lasting 1 min; and PURE\citep{pure}, which includes 59 videos from 10 subjects, with each clip lasting 1 min. For the PURE dataset, each subject performed six types of head movements: steady, talking, slow and fast translation between head movements and the camera plane, and small and medium head rotation. 

We evaluated the accuracy of four tasks, Heart Rate (HR), Standard Deviation of NN intervals (SDNN), Proportion of NN50 (pNN50), Root Mean Square of the Successive Differences (RMSSD). The HR was measured using the Welch method, coupled with a 30-210 BPM bandpass filter. The HRV analysis was conducted using the HeartPy toolkit\citep{9heartpy}, \citep{heartpy1}.

Each task utilized two metrics: Mean Absolute Error (MAE), Root Mean Square Error (RMSE).

\subsection{Intra-dataset Testing}
We randomly divide the RLAP dataset into training, validation, and testing sets according to the subjects. The partition details are provided in the datasets. 
All algorithms are tested on both the entire test set. In the RLAP dataset, due to the varying and longer video lengths, all HR tasks employ a 30-second moving window, while other tasks utilize the entire video. For UBFC\citep{ubfcrppg} and PURE\citep{pure} datasets, the entire videos are used for HR and HRV tasks. Test results are shown in Table \ref{resultrlap}.

In the intra-dataset testing, the performance of Seq-rPPG in the HR task is similar to that of the state-of-the-art baselines, where the MAE metric surpasses the state-of-the-art baselines, while the RMSE is slightly behind TS-CAN and PhysNet. In HRV-related tasks (SDNN, pNN50, RMSSD), Seq-rPPG exhibits significant advantages, outperforming all baselines and demonstrating substantial potential of this architecture for HRV tasks.

\begin{table*}
\caption{Intra-dataset testing on RLAP. \textbf{Bold}: The best result.}
\label{resultrlap}
\centering
\scalebox{1.2}{
\begin{threeparttable}
\begin{tabular}{lccccccccc}
  \toprule
  \multirow{2}*{\bf{Method}} & \multicolumn{2}{c}{\bf{HR}} & \multicolumn{2}{c}{\bf{SDNN}} & \multicolumn{2}{c}{\bf{pNN50}} & \multicolumn{2}{c}{\bf{RMSSD}} \\
  \cmidrule(lr){2-3}\cmidrule(lr){4-5}\cmidrule(lr){6-7}\cmidrule(lr){8-9}
  ~ & MAE↓ & RMSE↓ & MAE↓ & RMSE↓ & MAE↓ & RMSE↓ & MAE↓ & RMSE↓ \\
  \midrule
  \rowcolor{light-gray} Seq-rPPG & \bf 1.07 & 4.15 & \bf 12.7 & \bf 18.7 & \bf 0.137 & \bf 0.168 & \bf 22.9 & \bf 30.6 \\
  DeepPhys\citep{deepphys} & 1.52 & 4.40 & 61.1 & 71.1 & 0.367 & 0.396 & 92.5 & 103 \\
  TS-CAN\citep{mttscan} & 1.23 & \bf 3.59 & 43.0 & 56.1 & 0.267 & 0.304 & 65.1 & 80.3 \\
  PhysNet\citep{physnet} & 1.12 & 4.13 & 30.2 & 37.3 & 0.293 & 0.319 & 61.0 & 67.7\\
  PhysFormer\citep{physformer} & 1.56 & 6.28 & 22.8 & 28.1 & 0.267 & 0.296 & 48.7 & 54.7 \\
  \hline
  CHROM\citep{chrom} & 6.86 & 15.57 & 56.1 & 65.1 & 0.398 & 0.420 & 98.9 & 109 \\
  POS\citep{pos} & 4.25 & 12.06 & 78.0 & 83.1 & 0.502 & 0.518 & 142 & 149 \\
  ICA\citep{ICA1} & 6.05 & 13.3 & 77.3 & 82.8 & 0.505 & 0.524 & 136 & 145 \\
  \bottomrule
\end{tabular}
     \begin{tablenotes}[flushleft]
     \item \textbf{HR}: Heart Rate, \textbf{SDNN}: Standard Deviation of NN Intervals, \textbf{pNN50}: Percentage of NN50 Divisions, \textbf{RMSSD}: Root Mean Square of Successive Differences, \textbf{MAE}: Mean Absolute Error, \textbf{RMSE}: Root Mean Square Error.
     \end{tablenotes}
\end{threeparttable}
}
\end{table*}

\begin{table*}
\caption{Cross-dataset testing on UBFC-rPPG with comparison of different training sets. \textbf{Bold}: The best result.}
\label{resultubfc}
\centering
\scalebox{1.2}{
\begin{threeparttable}
\begin{tabular}{lccccccccc}
  \toprule
  \multirow{2}*{\bf{Method}} & \multirow{2}*{\bf{Training set}} & \multicolumn{2}{c}{\bf{HR}} & \multicolumn{2}{c}{\bf{SDNN}} & \multicolumn{2}{c}{\bf{pNN50}} & \multicolumn{2}{c}{\bf{RMSSD}} \\
  \cmidrule(lr){3-4}\cmidrule(lr){5-6}\cmidrule(lr){7-8}\cmidrule(lr){9-10}
  ~ & ~ & MAE↓ & RMSE↓ & MAE↓ & RMSE↓ & MAE↓ & RMSE↓ & MAE↓ & RMSE↓\\
  \midrule
  \rowcolor{light-gray} Seq-rPPG & \multirow{5}*{RLAP} & \bf 0.918 & \bf 1.42 & \bf 5.15 & \bf 9.04 & \bf 0.078 & \bf 0.131 & \bf 11.1 & \bf 18.2 \\
  DeepPhys\citep{deepphys} & ~ & 1.30 & 2.39 & 31.2 & 42.2 & 0.357 & 0.388 & 60.8 & 72.8 \\
  TS-CAN\citep{mttscan} & ~ & 1.11 & 1.51 & 25.5 & 30.9 & 0.349 & 0.373 & 54.3 & 62.1 \\
  PhysNet\citep{physnet} & ~ & 0.962 & 1.47 & 13.9 & 17.4 & 0.214 & 0.234 & 28.7 & 33.1 \\
  PhysFormer\citep{physformer} & ~ & 1.05 & 1.55 & 10.4 & 13.2 & 0.159 & 0.179 & 20.3 & 24.5\\
  \hline
  Seq-rPPG & \multirow{5}*{PURE} & 1.22 & 1.84 & 18.8 & 29.0 & 0.237 & 0.274 & 35.1 & 45.9 \\
  DeepPhys\citep{deepphys} & ~ & 1.97 & 5.09 & 51.0 & 60.9 & 0.487 & 0.514 & 90.4 & 101.4\\
  TS-CAN\citep{mttscan} & ~ & 1.17 & 1.71 & 38.1 & 48.5 & 0.423 & 0.446 & 70.2 & 79.5 \\
  PhysNet\citep{physnet} & ~ & 1.29 & 1.83 & 27.3 & 31.5 & 0.388 & 0.413 & 57.5 & 63.9 \\
  PhysFormer\citep{physformer} & ~ & 1.60 & 3.07 & 22.4 & 26.1 & 0.357 & 0.376 & 48.1 & 52.5 \\
  \hline
  CHROM\citep{chrom} & ~ & 6.10 & 19.6 & 22.7 & 29.0 & 0.336 & 0.359 & 49.0 & 59.2 \\
  POS\citep{pos} & ~ & 2.54 & 8.97 & 43.6 & 49.9 & 0.524 & 0.539 & 93.4 & 100 \\
  ICA\citep{ICA1} & ~ & 1.59 & 2.55 & 44.8 & 52.3 & 0.488 & 0.508 & 91.5 & 101 \\
  \bottomrule
\end{tabular}
     \begin{tablenotes}[flushleft]
     \item \textbf{HR}: Heart Rate, \textbf{SDNN}: Standard Deviation of NN Intervals, \textbf{pNN50}: Percentage of NN50 Divisions, \textbf{RMSSD}: Root Mean Square of Successive Differences, \textbf{MAE}: Mean Absolute Error, \textbf{RMSE}: Root Mean Square Error.
     \end{tablenotes}
\end{threeparttable}
}
\end{table*}

\begin{table*}
\caption{Cross-dataset testing on PURE with comparison of different training sets. \textbf{Bold}: The best result.}
\label{resultpure}
\centering
\scalebox{1.2}{
\begin{threeparttable}
\begin{tabular}{lccccccccc}
  \toprule
  \multirow{2}*{\bf{Method}} & \multirow{2}*{\bf{Training set}} & \multicolumn{2}{c}{\bf{HR}} & \multicolumn{2}{c}{\bf{SDNN}} & \multicolumn{2}{c}{\bf{pNN50}} & \multicolumn{2}{c}{\bf{RMSSD}} \\
  \cmidrule(lr){3-4}\cmidrule(lr){5-6}\cmidrule(lr){7-8}\cmidrule(lr){9-10}
  ~ & ~ & MAE↓ & RMSE↓ & MAE↓ & RMSE↓ & MAE↓ & RMSE↓ & MAE↓ & RMSE↓\\
  \midrule
  \rowcolor{light-gray} Seq-rPPG & \multirow{5}*{RLAP} & \bf 0.318 & \bf 0.597 & \bf 11.2 & \bf 19.6 & \bf 0.111 & \bf 0.138 & \bf 17.7 & \bf 26.1 \\
  DeepPhys\citep{deepphys} & ~ & 3.10 & 8.63 & 93.5 & 99.4 & 0.485 & 0.514 & 150 & 158 \\
  TS-CAN\citep{mttscan} & ~ & 2.19 & 6.69 & 74.8 & 86.2 & 0.404 & 0.442 & 120 & 136 \\
  PhysNet\citep{physnet} & ~ & 0.420 & 0.666 & 21.3 & 33.3 & 0.224 & 0.257 & 38.3 & 50.1 \\
  PhysFormer\citep{physformer} & ~ & 1.56 & 9.45 & 20.6 & 31.1 & 0.203 & 0.230 & 35.6 & 45.3\\
  \hline
  Seq-rPPG & \multirow{5}*{UBFC} & 21.5 & 35.0 & 83.2 & 92.0 & 0.394 & 0.443 & 116 & 130 \\
  DeepPhys\citep{deepphys} & ~ & 8.31 & 15.3 & 88.0 & 93.2 & 0.631 & 0.644 & 145 & 150\\
  TS-CAN\citep{mttscan} & ~ & 16.1 & 23.8 & 96.9 & 99.6 & 0.681 & 0.687 & 156 & 158 \\
  PhysNet\citep{physnet} & ~ & 8.82 & 19.2 & 59.6 & 66.7 & 0.385 & 0.422 & 96.5 & 107 \\
  PhysFormer\citep{physformer} & ~ & 16.3 & 28.2 & 70.5 & 75.5 & 0.461 & 0.495 & 109 & 114\\
  \hline
  CHROM\citep{chrom} & ~ & 2.76 & 13.3 & 47.4 & 64.6 & 0.276 & 0.327 & 73.4 & 95.5 \\
  POS\citep{pos} & ~ & 0.402 & 0.658 & 64.3 & 74.2 & 0.483 & 0.508 & 127 & 145 \\
  ICA\citep{ICA1} & ~ & 0.805 & 3.36 & 76.5 & 85.6 & 0.461 & 0.492 & 144 & 159 \\
  \bottomrule
\end{tabular}
     \begin{tablenotes}[flushleft]
     \item \textbf{HR}: Heart Rate, \textbf{SDNN}: Standard Deviation of NN Intervals, \textbf{pNN50}: Percentage of NN50 Divisions, \textbf{RMSSD}: Root Mean Square of Successive Differences, \textbf{MAE}: Mean Absolute Error, \textbf{RMSE}: Root Mean Square Error.
     \end{tablenotes}
\end{threeparttable}
}
\end{table*}

\subsection{Cross-dataset Testing}
We use UBFC\citep{ubfcrppg} and PURE\citep{pure} as test sets separately. When using UBFC as the testing set, the training sets are RLAP and PURE. When using PURE as the testing set, the training sets are RLAP and UBFC. Refer to Tables \ref{resultubfc} and \ref{resultpure} for details.

In cross-dataset testing, Seq-rPPG leads all other baselines in all tasks, whether on the UBFC-rPPG\citep{ubfcrppg} dataset or the PURE\citep{pure} dataset. This indicates that Seq-rPPG not only possesses superior HRV prediction capabilities but also demonstrates robust generalization and transferability. Additionally, we tested results using different training datasets. For instance, when tested on UBFC-rPPG, models trained on RLAP generally outperformed those trained on PURE. Conversely, when tested on PURE, models trained on RLAP showed better performance than those trained on UBFC-rPPG. This suggests that the RLAP dataset is of higher quality and possesses greater generalization abilities, making it a preferable training set.

\subsection{Computational Overhead}
We tested the average frame time of several algorithms on a typical mobile device: a Raspberry Pi 4B (CPU: Cortex-A72 4 cores). Compared with the best models PhysNet and TS-CAN, Seq-rPPG has a lower computational overhead (about 1/30 of theirs) while having similar or better performance. See Table \ref{overhead} for details. 


\begin{table}[H]
\centering
 \caption{Computational Overhead on Mobile CPUs}
 \label{overhead}
 \begin{threeparttable}
\begin{tabular}{lccc}
   \toprule
   Model & Resolution & Frame FLOPs (M) & Frame Time (ms) \\
   \midrule
   \rowcolor{light-gray} Seq-rPPG & 8x8 & 0.26 & 0.36 \\
   DeepPhys\citep{deepphys} & 36x36 & 52.16 & 9.09 \\
   TS-CAN\citep{mttscan} & 36x36 & 52.16 & 10.72 \\
   PhysNet\citep{physnet} & 32x32 & 54.26 & 9.69 \\
   PhysFormer\citep{physformer} & 128x128 & 323.80 & 150 \\
   \bottomrule
\end{tabular}
 \end{threeparttable}
 \end{table}

\vspace{-0.1cm}
\section{Discussion}

\subsection{RLAP Dataset}

Tables \ref{resultubfc} and \ref{resultpure} indicate that the Seq-rPPG model and all baseline models perform better when trained on the RLAP dataset, especially in the tasks on HRV, corroborating the stance we advocated during our data collection process. Contrasting with other datasets, such as the UBFC-rPPG\citep{ubfcrppg} dataset, which is relatively simplistic and lacks complex scenarios, and where the BVP labels are not perfectly synchronized, this simplicity leads to reduced generalizability, and the offset in BVP labels affects training for HR and HRV tasks. Thus, it is unsuitable for use as a training set, as models trained on UBFC-rPPG demonstrate poor performance. The results are significantly better on the PURE\citep{pure} dataset, which consists of six diverse scenarios and where BVP signals are as synchronously matched with the visuals as possible, validating our hypothesis regarding the importance of dataset collection synchronization. However, our RLAP dataset achieves the best performance; in comparison to PURE, RLAP not only contains complex scenarios and synchronized labels but also includes various emotional and engagement tasks that induce HRV changes, along with a substantially larger data volume — a total of 32.7 hours of video, whereas PURE comprises merely one hour.

\subsection{Seq-rPPG Method}

In the intra-dataset and cross-dataset testing, Seq-rPPG demonstrates excellent performance in tasks related to HR and HRV. This performance is attributed to two key features of Seq-rPPG: its large context window and a unique time-frequency multi-layer one-dimensional decoder architecture. Typically, rPPG algorithms consider joint spatio-temporal modeling, wherein high-resolution multi-frame images are inputted, modeling across both temporal and spatial dimensions. However, due to computational constraints, these models usually cannot afford long-range temporal associations, as evidenced by their relatively small time windows, such as the 160 frames in PhysFormer\citep{physformer} and the 128 frames in PhysNet\citep{physnet}. Extending these temporal associations in such spatially-focused models would significantly increase computational load, with time complexity of $O(N^3)$. However, Seq-rPPG excels in this regard. Unlike some recent approaches focused on spatial\citep{tang2024camera}, \citep{liu2024spiking}, \citep{liu2022mobilephys}, by encoding video input into a one-dimensional signal through a straightforward encoder, Seq-rPPG allows for an increased time window even under limited computational resources, bringing the complexity down to $O(N)$. Additionally, its fast Fourier transform convolution layer—also known as the spectral transformation layer—enables the construction of a global receptive field, meaning that the effective time window equals the length of the model’s input. Collectively, Seq-rPPG maximally extends the temporal window to establish long-range associations, better capturing periodic features and hence enhancing performance.

\begin{figure}[!htb] 
\centering 
\includegraphics[width=0.5\textwidth]{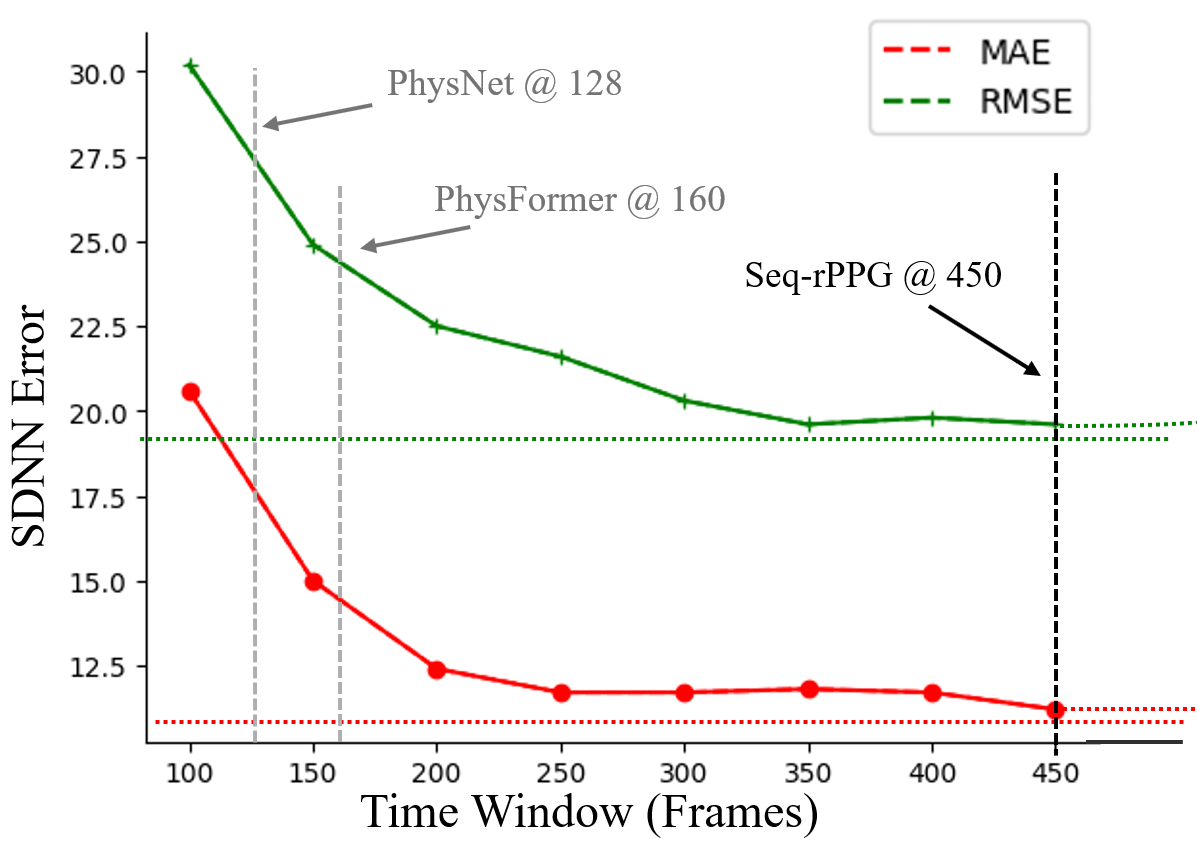} 
\caption{The relationship between the time window size of Seq-rPPG and the SDNN error.
} 
\label{timewindow} 
\end{figure}

To further validate the sensitivity of the rPPG algorithm to time windows, we configured various parameters for Seq-rPPG, ranging from a 100 frames time window to a maximum of 450 frames. We trained on RLAP and tested on PURE, observing the SDNN error curve as the time window varied, as shown in Fig. \ref{timewindow}. At first, the error decreases rapidly, but as the window size continues to increase beyond 400 frames, the performance enhancement becomes less significant. Consequently, we opted for a window size of 450 frames for Seq-rPPG, corresponding to a 15-second video input. For reference, the time windows of PhysNet\citep{physnet} and PhysFormer\citep{physformer} are also marked in the graph, which explains why Seq-rPPG exhibits stronger performance on the HRV tasks.


\begin{figure}[!htb] 
\centering 
\includegraphics[width=0.5\textwidth]{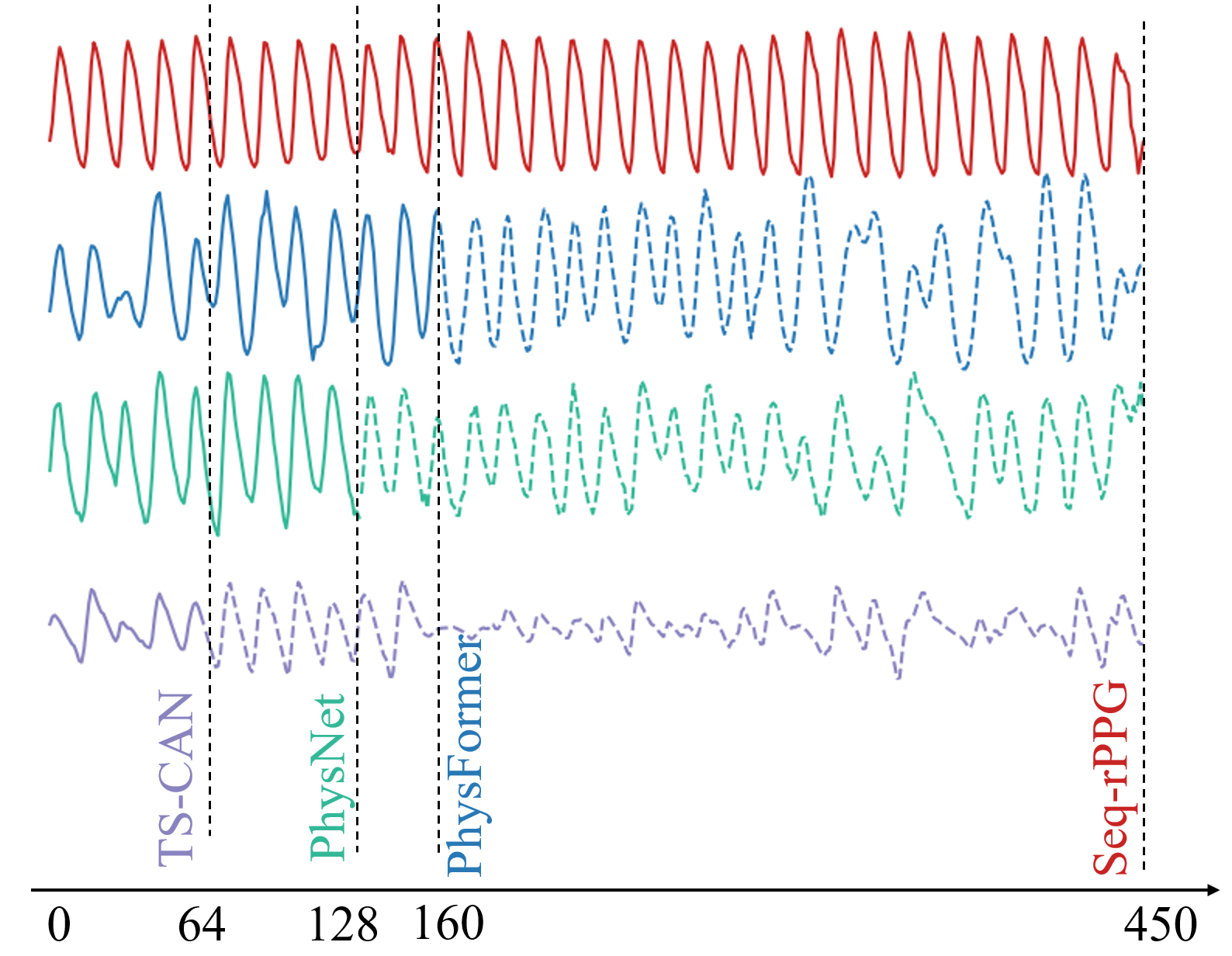} 
\caption{The BVP waveforms output by four models during head movements, where the solid lines indicate the size of the time window. 
} 
\vspace{-0.2cm}
\label{snr} 
\end{figure}

As illustrated in Fig. \ref{snr}, within the PURE dataset, the presence of partial head movements often leads to distortions in model outputs. By plotting the waveform of the model output, it is observed that smaller time windows lead to a diminished capability to cope with distortions, resulting in outputs that contain more substantial noise. The Seq-rPPG model employs a large window of 450 frames, indicating its robust ability to mend noise generated by head movements, and its decoder architecture is capable of generating authentic BVP waveforms.

Due to the one-dimensional structure of Seq-rPPG, it obtains large time windows at a low computational cost. It has significant advantages in terms of accuracy while having very little computational overhead, only 3\% of other baselines as shown in Table \ref{overhead}. Therefore, we believe it is a highly potential structure that can achieve high accuracy while being lightweight.

\section{Conclusion}

In this study, we collected the RLAP public dataset, which is suitable for remote learning and affective computing and includes the BVP signal designed for HRV tasks. In past rPPG datasets, many studies did not focus on the strict synchronization between labels and videos, which led to pulse peak shifts due to frame rate fluctuations and signal delays, posing significant challenges for HRV tasks. RLAP addresses this issue and also publicized its data collection tool, PhysRecorder, aiding in the collection of more high-quality datasets in the future.

We proposed the Seq-rPPG algorithm, which performed excellently on HRV tasks. Through analysis, it was demonstrated that the time window and temporal receptive field of the algorithm were crucial for HRV tasks, which guided the design of future rPPG algorithms. Seq-rPPG was also a lightweight algorithm that could easily run in real-time on mobile devices, facilitating the widespread application of rPPG algorithms.
\section*{ACKNOWLEDGMENT}
This work is supported by the foundation of the National Key Laboratory of Human Factors Engineering, Grant NO. HENKL.2024W06, the Natural Science Foundation of China (NSFC) under Grant No. 62366043, 62472244, Tsinghua University Initiative Scientific Research Program, Beijing Natural Science Foundation(No.QY23124, No.QY24248). National Natural Science Foundation of China under Grant 62277029, the Humanities and Social Sciences of China MOE under Grants 20YJC880100. The experimental procedures involving human subjects described in this paper were approved by the Institutional Review Board. 
\bibliographystyle{plain}
\bibliography{IEEE-conference-template-062824}
\end{document}